\documentclass[journal]{IEEEtran} 

\IEEEoverridecommandlockouts

\usepackage{graphicx,wrapfig,fullpage,amsmath,hhline,epsfig,verbatim,url,amssymb,multicol,multirow,cite}
\usepackage{times,color,soul} 
\usepackage[left=0.75in,top=0.70in,right=0.75in,bottom=0.80in]{geometry}
\usepackage{amsbsy,latexsym,mathrsfs,mathtools}

\usepackage{mathptmx} 
\DeclareMathAlphabet{\mathcal}{OMS}{cmsy}{m}{n}
\usepackage{bm}
\usepackage{float}
\usepackage{color,soul}
\usepackage{dsfont}
\usepackage{comment}
\usepackage{algorithm}
\usepackage{algorithmic}
\usepackage{psfrag}   

\usepackage{url}
\usepackage{hyperref}
\usepackage{setspace}
\usepackage[table]{xcolor}
\usepackage{paralist}
\usepackage{booktabs}
\usepackage[switch]{lineno}

\usepackage{balance}
\usepackage{tabularx}
\usepackage{tikz}

\newtheorem{thm}{Theorem}
\newtheorem{rem}{Remark}

\DeclareUnicodeCharacter{FF0C}{,}

\begin{document}

\title{\LARGE \bf {HT-LIP Model based Robust Control of Quadrupedal Robot Locomotion under Unknown Vertical Ground Motion}}

\author{Amir~Iqbal,~
Sushant~Veer,~Christopher~Niezrecki,~and~
        Yan~Gu
\thanks{
This work was supported by the National Science Foundation under Grants 1934280 and 2046562 and by the Office of Naval Research under Grant N00014-21-1-2582.
A. Iqbal and C. Niezrecki are with the Department of Mechanical Engineering, University of Massachusetts Lowell, Lowell, MA 01854, U.S.A.
S. Veer is with NVIDIA Research, Santa Clara, CA 95051, U.S.A.
Y. Gu is with the School of Mechanical Engineering, Purdue University, West Lafayette, IN 47907, U.S.A.
Corresponding author: Y. Gu ({\tt\small yangu@purdue.edu}).}}

\maketitle

\begin{abstract}
This paper presents a hierarchical control framework that enables robust quadrupedal locomotion on a dynamic rigid surface (DRS) with general and unknown vertical motions.
The key novelty of the framework lies in its higher layer, which is a discrete-time, provably stabilizing footstep controller.
The basis of the footstep controller is
a new hybrid, time-varying, linear inverted pendulum (HT-LIP) model that is low-dimensional and accurately captures the essential robot dynamics during DRS locomotion.
A new set of sufficient stability conditions are then derived to directly guide the controller design for ensuring the asymptotic stability of the HT-LIP model under general, unknown, vertical DRS motions.
Further, the footstep controller is cast as a computationally efficient quadratic program that incorporates the proposed HT-LIP model and stability conditions.
The middle layer takes the desired footstep locations generated by the higher layer as input to produce kinematically feasible full-body reference trajectories, which are then accurately tracked by a lower-layer torque controller.
Hardware experiments on a Unitree Go1 quadrupedal robot confirm the robustness of the proposed framework under various unknown, aperiodic, vertical DRS motions and uncertainties (e.g., slippery and uneven surfaces, solid and liquid loads, and sudden pushes).

\end{abstract}

\IEEEpeerreviewmaketitle


\vspace{-0.15in}

\section{Introduction}
\label{Section-Introduction}
\vspace{-0.05 in}

Due to the prevalence of uncertainties in real-world environments, robustness is a crucial performance measure of legged robot control.
Various control approaches~\cite{cheetah3,hutter2017anymal_RW_Application,RAL_2020quadrupedal_hamed,hwangbo2019learning} have achieved remarkably robust locomotion in a wide variety of unstructured, {\it static} environments (e.g., sand, grass, hiking trails, and creeks).
Yet, since the previous approaches typically assume a static ground, they may not be effective for a dynamic rigid surface (DRS), which is a rigid surface moving in the inertial frame and can persistently and continuously perturb the robot movement.
This paper introduces a reduced-order model based control framework that achieves robust quadrupedal trotting on a DRS with a general and unknown vertical motion.

\vspace{-0.15in}

\subsection{Related Work}
\vspace{-0.05 in}

\subsubsection{Related work on DRS locomotion control}
Recently, there has been a growing interest in addressing the problem of DRS locomotion control~\cite{Henze_Passivity,Englsberger2018DCMBasedGG,Yamenzheng2011ball,iqbal2020provably,BallMan2020Koshil_yang,asano2021modeling}. 
Henze et al. \cite{Henze_Passivity} have proposed a passivity-based controller based on a full-order robot model for humanoid balancing on a rigid rocker board.
Englsberger et al. \cite{Englsberger2018DCMBasedGG} have proposed a walking gait generator
for humanoid walking on a rigid surface with a constant linear velocity.
However, these studies do not address surfaces with notable, varying accelerations.

Researchers have also explored locomotion control for floating-base, rigid platforms with inertia comparable to the robot, including rolling rigid balls~\cite{Yamenzheng2011ball,BallMan2020Koshil_yang} and floating islands~\cite{asano2021modeling}. 
Still, the robot control problem for rigidly actuated or heavyweight
DRSes (e.g., trains, vessels, and airplanes), whose dynamics are barely affected by the physical robot-surface interaction, remains under-explored.
Our previous legged robot controllers for DRS locomotion~\cite{iqbal2020provably,iqbal2021extended,iqbal2022drs,gao2022invariant,gao2022time} have focused on such surfaces.
Yet, since they assume a periodic (and even sinusoidal) surface motion whose entire time profile is accurately known ahead of time,
they cannot address unknown or aperiodic DRS motions.

\begin{figure*}[t]
    \centering
    \includegraphics[width= 0.9\linewidth]{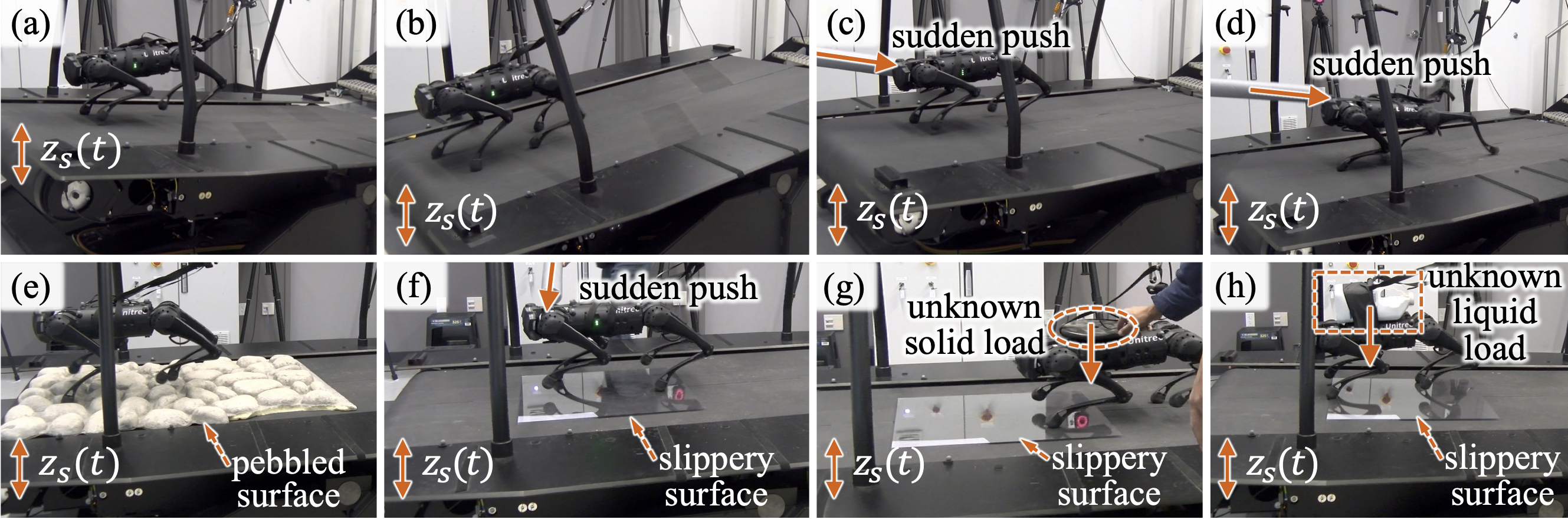}
    \vspace{-0.1 in}
    \caption{Snapshots of experiments. All experiments are under the  unknown and aperiodic vertical surface motion $z_s(t)$ as shown in (a) and (b). The robot also experiences additional unknown disturbances, which include: (c) sudden pushes that result in (d) an irregular robot posture just after a push; (e) rocky surface with a peak height of 10 cm; (f) smooth glass surface; (g) solid load (36\% of the robot's mass); and (h) liquid load (32\% of the robot's mass).} 
    \label{Fig:HardwareValidation}
    \vspace{-0.2 in}
\end{figure*}

\subsubsection{Related work on reduced-order models}
Reduced-order models describe the robot's essential dynamics.
By considering the relatively simple reduced-order models instead of the complex full-order models, motion generators can more efficiently plan desired trajectories, enabling
quick reaction to disturbances for robust locomotion.

One widely used reduced-order model is the linear inverted pendulum (LIP) model \cite{kajita20013d}.
Thanks to its linearity, low dimensionality, and analytical tractability, the LIP has served as a basis for the closed-form analysis, online motion generation, and real-time control of bipedal \cite{kajita20013d,kajita2010biped,pratt2006capture} and quadrupedal \cite{mastalli2020motion} locomotion on static surfaces. 
The classical LIP describes a legged robot as a point mass, which corresponds to the robot's center of mass (CoM), atop a massless leg, with the point foot located at the robot's center of pressure (CoP)~\cite{kajita20013d,caron2019capturability, LIP_varyingHeight_caron2020biped}.

The classical LIP model has been expanded to capture the hybrid dynamics of legged locomotion on a stationary surface~\cite{xiong2021_3d,gong2020angular,dai2022bipedal,paredes2022resolved}, which include continuous leg-swinging dynamics and discrete foot-switching behaviors.
Using the theory of linear, hybrid, time-invariant systems, the asymptotic stability condition for the hybrid LIP (H-LIP) model under a discrete-time footstep controller has been constructed to enable robust locomotion under external pushes~\cite{xiong2021_3d,gong2020angular}.
Yet, the model and stability condition may not be valid under a significant DRS motion since they assume a static ground.

Although our recent study on quadrupedal walking has analytically extended the continuous-time LIP model~\cite{kajita20013d} from static to dynamic surfaces~\cite{iqbal2022drs,iqbal2021extended},
the modeling and analysis do not consider hybrid robot dynamics.

\vspace{-0.15 in}
\subsection{Contributions}
\vspace{-0.05 in}

This paper introduces a reduced-order model based control approach that achieves robust quadrupedal trotting on rigidly actuated or heavyweight DRSes with aperiodic and unknown
vertical motions (e.g., ships and airplanes).
Some of the analytical results reported in this paper have been previously presented in \cite{iqbal2023asymptotic}, which are the derivation of the proposed HT-LIP model and the preliminary stability analysis of the HT-LIP.
This study makes the following new, substantial contributions:
(a) generalization of the stability condition in~\cite{iqbal2023asymptotic} to enlarge the solution space for controller design under unknown DRS motions;
(b) formulation of a robust footstep controller as a computationally efficient quadratic program that enforces stability conditions even under unknown, vertical motions;
(c) derivation of a hierarchical control approach that incorporates the proposed quadratic program;
(d) stability analysis for the full-order model under the proposed control approach; and 
(e) experimental validation under various uncertainties~(Fig. \ref{Fig:HardwareValidation}).

\vspace{-0.1 in}
\section{Stabilization of a Hybrid Time-Varying LIP}
\vspace{-0.05 in}
\label{Section-Dyanmics}

This section introduces a reduced-order model that captures the essential hybrid robot dynamics associated with quadrupedal trotting on a DRS with a general vertical motion, along with its stabilizing control law.

\vspace{-0.15in}
\subsection{Open-Loop Reduced-Order Model}
\vspace{-0.05 in}

To derive the proposed reduced-order model, we extend the classical H-LIP model \cite{xiong2021_3d} from static surfaces to DRSes by combining the H-LIP and our previous continuous-phase time-varying LIP model~\cite{iqbal2022drs} derived for DRSes.
The resulting model, as illustrated in Fig.~\ref{Fig:LIPM_sketch}, is a {\it hybrid}, {\it time-varying} LIP model, which we call ``HT-LIP''.

\subsubsection{Model assumptions}
The proposed model derivation considers the following simplifying assumptions: 
\begin{itemize}
\item [(A1)] The absolute vertical acceleration of the DRS is bounded
and is locally Lipschitz in time.
\item [(A2)] The desired duration of the continuous phase during the HT-LIP stepping is bounded for all walking steps.
\item [(A3)] The CoM maintains a constant height above the CoP (i.e., the support point {\small $S$} in Fig. \ref{Fig:LIPM_sketch}).
\end{itemize}

Assumption (A1) holds for common real-world dynamic platforms since their acceleration is
continuous and bounded and does not change abruptly~\cite{bergdahl2009wave_ShipMotionRange}.
Assumption (A2) is reasonable as it ensures a finite duration for each continuous phase of the HT-LIP and prevents Zeno behavior~\cite{zhang2001zeno}.
Assumption (A3) helps avoid kinematic singularity induced by an overly stretched knee joint, and  ensures the linearity of an inverted pendulum model~\cite{kajita20013d} as explained later.

\subsubsection{Continuous phases}
Under assumption (A3), the continuous-phase dynamics of a 3-D inverted pendulum model along the $x$- and $y$-axes of the world frame are 
linear and share the same form,
as explained in our previous work on continuous-time LIP modeling for DRSes~\cite{iqbal2023asymptotic}.
Without loss of generality and for brevity, the subsequent analysis considers the HT-LIP model in the {\small $x$}-direction (see Fig.~\ref{Fig:LIPM_sketch}).

We use {\small$\ddot{z}_{s}(t)$}, {\small $g$}, and {\small $z_0$} to respectively denote the vertical acceleration of the support point $S$, the magnitude of gravitational acceleration, and the CoM height above {\small $S$}.
Here, the time argument {\small $t$} is kept in the notation of the surface acceleration {\small $\ddot{z}_{s}(t)$} to highlight its explicit time dependence.

Denoting the horizontal CoM position relative to point {\small $S$} as {\small $x$}, we express the continuous-phase equation of motion for the HT-LIP in the {\small $x$}-direction as the following continuous-time, time-varying, linear, homogeneous system:
\vspace{-0.05in}
\begin{equation} 
\small
\ddot{x} = \frac{\ddot{z}_{s}(t) + g}{z_0} x.
\label{Eq_xsc} 
\vspace{-0.02in}
\end{equation}

\subsubsection{Discrete foot switching}
Besides continuous dynamics, the proposed HT-LIP also considers the discrete foot-landing event when the stance and support feet switch roles.
We use {\small $\tau_n$} to denote the {\small $n^{\text{th}}$} switching instant with {\small $n \in \mathbb{N}$}.
Further, we denote the time instant just before and after the {\small $n^{\text{th}}$} switching instant as {\small $\tau_n^- $} and {\small $\tau_n^+ $}, respectively.
For notational brevity, we introduce {\small $\star|_n^-:=\star(\tau_{n}^-)$} and {\small $\star|_n^+:=\star(\tau_{n}^+)$}.

\begin{figure}[tb]
    \centering
    \includegraphics[width= 1\linewidth]{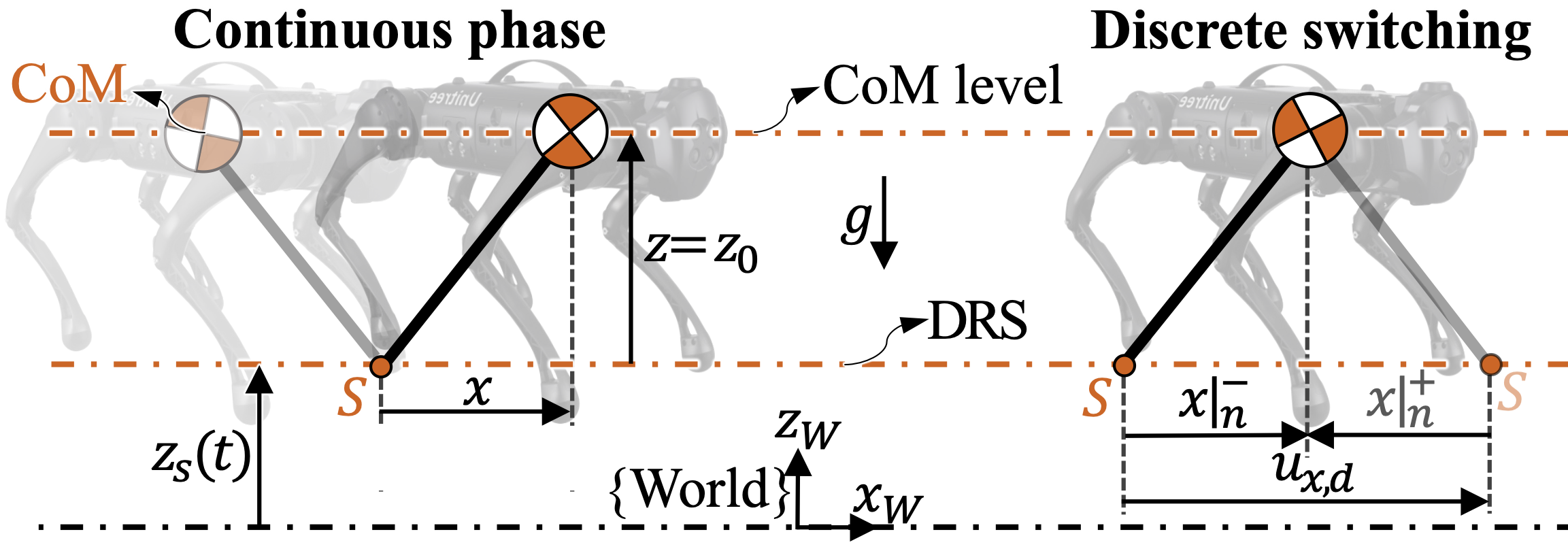}
    \vspace{-0.25 in}
    \caption{An illustration of the proposed HT-LIP model in the sagittal plane. The model describes the time-varying dynamics of the point mass (located at the CoM) under the vertical DRS displacement $z_s(t)$. It also captures the hybrid nature of legged locomotion, including both the continuous foot-swinging phase and the discrete foot-switching behavior.}
    \label{Fig:LIPM_sketch}
        \vspace{-0.2 in}
\end{figure}

At the switching timing, the location of the support point {\small $S$} on the DRS is reset, resulting in an sudden jump in the relative CoM position {\small $x$}.
As illustrated in Fig.~\ref{Fig:LIPM_sketch}, the relative CoM position just after the switching, {\small $x|_n^+$}, is given by:
\vspace{-0.05in}
\begin{equation}
\small
    x|_n^+ =x|_n^- - u_{x,d},
    \label{Eq_CoM_Pos_Jump}
    \vspace{-0.03in}
\end{equation}
where {\small $u_{x,d}$} is the new support-foot position relative to the previous one in the {\small $x$}-direction.

The CoM velocity stays continuous at the switching instant, that is, {\small $\dot{x}|_n^+=\dot{x}|_n^-$}, 
because the angular momentum of the CoM about the contact point {\small $S$} is conserved and the CoM height remains constant above {\small $S$} within continuous phases (i.e., assumption (A3))~\cite{gong2020angular}.

Combining the continuous dynamics in \eqref{Eq_xsc} and the discrete jump in \eqref{Eq_CoM_Pos_Jump} yields the proposed HT-LIP model as:
\vspace{-0.05in}
\begin{equation}
\small
\begin{cases}
\mathbf{\dot{X}}= \boldsymbol{\alpha}(t) \mathbf{X}
&\text{if}~t\neq \tau_n^-, \\
\mathbf{X}(\tau_n^+)=\mathbf{X}(\tau_n^-) + \boldsymbol{\beta} u_{x,d}~~~~&\text{if}~ t= \tau_n^-,
\label{Eq-HybridSys}
\end{cases}
\vspace{-0.03in}
\end{equation}
where {\small $\mathbf{X} := [x ,~\dot{x} ]^T$}
and {\small $\boldsymbol{\beta}:=[-1,~0]^T$}.
The matrix {\small $\boldsymbol{\alpha}(t)$} is defined as
{\small$\boldsymbol{\alpha}(t) := { \begin{bmatrix}
0 & 1 \\
f(t) & 0
\end{bmatrix}
}
$}
with {\small $f(t) := \frac{\ddot{z}_{s}(t) + g}{z_0}$}.
Similar to {\small $z_{s}(t)$}, we keep the time argument {\small $t$} in the notation of {\small $f(t)$} and {\small $\boldsymbol{\alpha}(t)$} to highlight their explicit time dependence.

\subsubsection{Open-loop step-to-step (S2S) model}
The S2S model of the HT-LIP compactly describes the hybrid evolution of the HT-LIP during a gait cycle, which is used to construct the proposed stability conditions of the HT-LIP later.

Integrating the continuous dynamics and iterating the discrete jump map based on \eqref{Eq-HybridSys} yields the S2S model as:
\vspace{-0.05in}
\begin{equation}
\small
\begin{aligned}
\mathbf{X}|_{n+1}^-
&= 
\boldsymbol{\Phi}(f(t);\tau_{n+1}^-,\tau_{n}^+)
(\mathbf{X}|_{n}^- + \boldsymbol{\beta} u_{x,d}),
\label{EQ_S2S_HLIP_2}
\end{aligned}
\vspace{-0.03in}
\end{equation}
where {\small $\boldsymbol{\Phi}(f(t);\tau_{n+1}^-,\tau_{n}^+) :=\int_{\tau_n^+}^{\tau_{n+1}^-} \text{exp} \big({{\boldsymbol{\alpha}}(t)} \big) dt$} is the state-transition matrix of the {\small $n^{th}$} continuous phase from {\small $\tau_{n}^+$} to {\small $\tau_{n+1}^-$}. Here {\small $\text{exp} (\cdot)$} is a matrix exponential function.

\vspace{-0.15 in}
\subsection{Discrete Footstep Control for HT-LIP}
\vspace{-0.05 in}

\label{Sec_Foot_Stepping_Law}

While the continuous-time portion of the HT-LIP model is unstable~\cite{iqbal2021extended} and uncontrolled as indicated by \eqref{Eq-HybridSys}, 
the discrete-time footstep behavior is directly commanded by the foot displacement {\small $u_{x,d}$}.
Thus, 
we design a discrete-time footstep control law based on the HT-LIP model that aims to asymptotically stabilize the desired state trajectory, denoted as {\small $\mathbf{X}_{r}(t)$}; i.e., to drive the state trajectory {\small $\mathbf{X}(t)$} to track the desired trajectory {\small $\mathbf{X}_{r}(t)$} as time goes to infinity.

The tracking error is defined as {\small $\mathbf{e}:=\mathbf{X}-\mathbf{X}_{r} =: [e,~ \dot{e}]^T$}, where {\small $x_r$} and {\small $\dot{x}_r$} are the elements of {\small $\mathbf{X}_r$}, i.e., {\small $\mathbf{X}_r=[x_r,\dot{x}_r]^T$}.

By incorporating the error {\small $\mathbf{e}$}, the discrete HT-LIP stepping controller {\small $u_{x,d}$} at the switching instant {\small $\tau_n^-$} is designed as:
\vspace{-0.05in}
\begin{equation}
\small
u_{x,d}= {u}_{x,r}+\mathbf{K}\mathbf{e}|_n^-.
\label{Eq-FeedbackSteppingControl}
\vspace{-0.03in}
\end{equation}
Here {\small ${u}_{x,r}:=x_r|_n^- -x_r|_n^+$} is the desired foot-landing position of the desired trajectory {\small $\mathbf{X}_r(t)$}, and {\small $\mathbf{K}  := [k_{1},~k_{2}]$}
is the feedback gain to be designed later for asymptotic stabilization of {\small $\mathbf{X}_r(t)$}.

From the feedback control law \eqref{Eq-FeedbackSteppingControl} and the open-loop S2S dynamics \eqref{EQ_S2S_HLIP_2},
the closed-loop S2S error dynamics become:
\vspace{-0.05in}
\begin{equation}
\small
   \mathbf{e}|_{n+1}^-= \mathbf{A}_{d,n}\mathbf{e}|_{n}^-,
   \label{Eq-S2SDynamics_final}
\vspace{-0.03in}
\end{equation}
where {\small $\mathbf{A}_{d,n}$} is the S2S error state-transition matrix and is defined as
 {\small$\mathbf{A}_{d,n}:=\boldsymbol{\Phi}(f(t);\tau_{n+1}^-,\tau_{n}^+)( \mathbf{I}+ \boldsymbol{\beta}\mathbf{K})$} with {\small $\mathbf{I}$} an identity matrix with an appropriate dimension.

\begin{figure*}[tb]
    \centering
    \includegraphics[width= 1\linewidth]{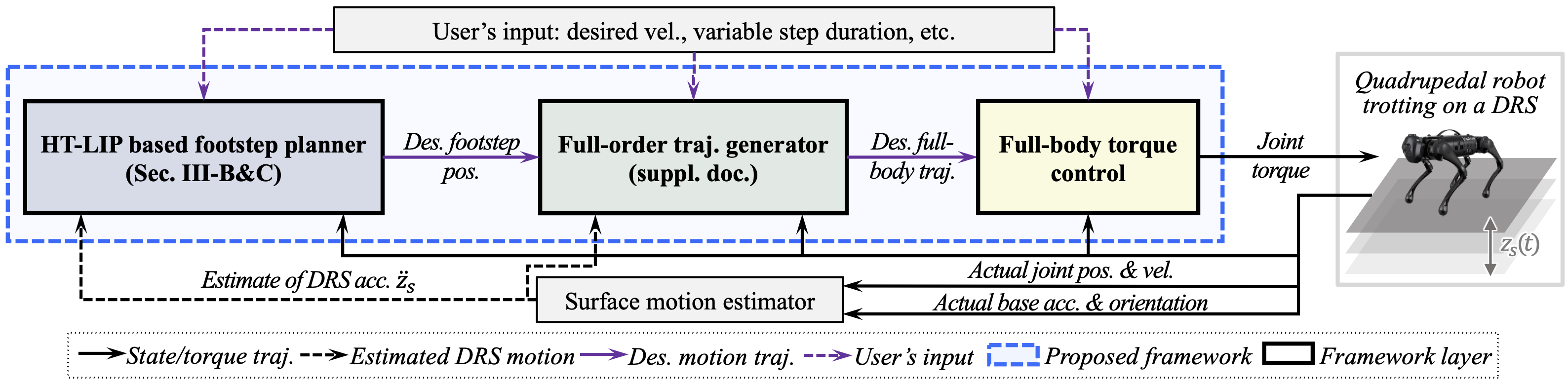}
    \vspace{-0.3in}
    \caption{Illustration of the proposed hierarchical control framework. The higher layer generates the desired footstep locations. The middle layer employs a full-order kinematics model to plan physically feasible full-body trajectories. The lower-layer controller tracks the desired full-body trajectories.}
    \vspace{-0.2in}
    \label{Fig:BlockDiagram_OAF}
\end{figure*}

\vspace{-0.15 in}
\section{HT-LIP Based Footstep Planning} 
\label{Section-QuadDesiredTraj}
\label{Section-FootStepControlGain}
This section presents the overall structure and higher-layer footstep planner of the proposed hierarchical control framework.
The framework aims to achieve robust quadrupedal trotting on a DRS with an unknown vertical motion.

One effective way to realize robust locomotion is to plan the physically feasible footstep locations in real-time~\cite{xiong2021_3d,gong2020angular}.
However, realizing online footstep planning is substantially challenging due to the complex robot dynamics, which are hybrid, nonlinear, time-varying, and high-dimensional.

Another challenge in achieving robust locomotion is the underactuation associated with quadrupedal trotting. 
With 13 DoFs and 12 independently actuated joints, a typical quadrupedal robot (e.g., Unitree's Go1) has one degree of underactuation during trotting and accordingly two-dimensional unactuated dynamics.
Realizing robust locomotion under underactuation is complex because (a) while the directly actuated portion of the actual robot dynamics can be well regulated, the unactuated subsystem may not be directly altered by joint torque commands~\cite{gong2020angular} and (b) as indicated in our prior analysis~\cite{iqbal2022drs}, the unactuated system during continuous phases is inherently unstable under real-world DRS motions (e.g., ship motions in sea waves).

To achieve robust locomotion, the proposed framework adopts the classical hierarchical framework structure~\cite{xiong2021_3d,gong2020angular} and contains three layers.
The key novelty of the framework lies in its higher-layer footstep planner, which is presented in this section.
The middle layer and the stability analysis of the complete closed-loop unactuated subsystem are respectively given in Secs. III and IV of the supplementary file. Details of the lower layer are omitted since the existing torque controller~\cite{cheetah3} is adopted.

\vspace{-0.15 in}
\subsection{Framework Structure}
\vspace{-0.05 in}

\subsubsection{Higher-layer footstep planning}
To reject uncertainties for ensuring robust locomotion, the proposed higher layer efficiently generates the desired, physically feasible footstep locations and CoM position trajectories in real-time.

To guarantee the planner's feasibility, we use the proposed HT-LIP model to approximate the robot dynamics in the higher-layer planning.
The HT-LIP model is reasonably accurate because today's legged robots typically have heavy trunks and lightweight limbs, thus closely emulating an inverted pendulum~\cite{iqbal2022drs}.
Meanwhile, thanks to its linearity and low dimension, using the HT-LIP model can also ensure planning efficiency for real-time motion generation.

Further, to ensure the stability of the hybrid, time-varying, nonlinear, and underactuated robot dynamics, we construct the higher-layer planner as a real-time footstep controller of the HT-LIP, which indirectly stabilizes the unactuated dynamics by provably stabilizing the HT-LIP.
This footstep controller is the key novelty of the higher-layer planner, and is introduced in subsections B and C.


\subsubsection{Middle-layer full-body trajectory generation}
Based on the robot's full-order kinematics model, the middle layer efficiently translates the output from the higher layer (i.e., the desired footstep location and CoM trajectories) into the desired full-body trajectories.
The translation also agrees with assumptions (A1)-(A3) underlying the HT-LIP model, further reducing the discrepancy between the actual robot dynamics and the model for planning feasibility.

\subsubsection{Lower-layer full-body control}
Considering its high performance in ensuring gait feasibility and motion tracking accuracy, the lower layer adopts the existing controller~\cite{cheetah3} that outputs the joint torque to track the desired full-body trajectories based on a single rigid body model. 
Both the middle and lower layers approximate the robot's CoM at the base/trunk center.

\vspace{-0.15 in}
\subsection{Stability Condition under Unknown DRS Motions}
\vspace{-0.05 in}

The design of the proposed higher-layer footstep planner begins with the construction of the asymptotic stability condition of the HT-LIP model under unknown DRS motions.

\subsubsection{Supreme model of HT-LIP}

The proposed asymptotic stability condition is built on a supreme model of the S2S error dynamics in~\eqref{Eq-S2SDynamics_final}, which is derived next. 

By definition, the function {\small $f(t)$} is both positive and bounded for {\small $t \in \mathbb{R}^+$} and locally Lipschitz under the assumption (A1).
We use {\small $\overline{f}_n$} to represent any positive constant parameter no less than the supremum of {\small $f(t)$} over {\small $t \in (\tau_n,\tau_{n+1}]$} (i.e., {\small $\overline{f}_n$} should satisfy {\small $\overline{f}_n \geq {\text{sup} f(t)}$} on {\small $t \in (\tau_n,\tau_{n+1}]$}.

Since the continuous-phase error system is {\small $\ddot{{e}} = {f}(t) {e}$},
we define its supreme model as:
\vspace{-0.03in}
\begin{equation}
\small
    \ddot{\overline{e}} = \overline{f}_n \overline{e},
    \vspace{-0.03 in}
        \label{eq: supreme}
\end{equation}
where {\small $\overline{e}$} is the solution of this model.
Because the supremum model is linear and time-invariant, 
its state-transition matrix, denoted as {\small $\overline{\boldsymbol{\Phi}}$}, satisfies
{\small$\overline{\boldsymbol{\Phi}}(\overline{f}_n;\tau_{n+1}^-,\tau_n^+) =\overline{\boldsymbol{\Phi}}(\overline{f}_n;\Delta \tau_{n+1},0)$}, where {\small $\Delta \tau_{n+1}:= \tau_{n+1}^- - \tau_n^+$} denotes the duration of the {\small $n^{\text{th}}$} continuous phase.
Accordingly, the S2S state-transition matrix of the supreme model is defined as 
\vspace{-0.03in}
\begin{equation}
    \small
    \overline{\mathbf{A}}_{d,n}:= \overline{\boldsymbol{\Phi}}(\overline{f}_n;\Delta \tau_{n+1},0)(\mathbf{I} +\boldsymbol{\beta}\mathbf{K}).
    \vspace{-0.03in}
\end{equation}


\subsubsection{Asymptotic stability condition on S2S dynamics}
We first introduce the sufficient condition for the asymptotic stability of the closed-loop S2S error model in \eqref{Eq-S2SDynamics_final}. 

\begin{thm}[\textbf{Sufficient stability condition on S2S dynamics}]
{\it Consider assumptions (A1) and (A2).
Define
\vspace{-0.03in}
\begin{equation}
\small
a_{d,n}
:=
\|\overline{\mathbf{A}}_{d,n} \|_\infty,
\vspace{-0.03in}
\end{equation}
where {\small $ \| \star \|_\infty$} is the infinity norm of the matrix {\small $\star$}.
The closed-loop S2S error dynamics in \eqref{Eq-S2SDynamics_final} is globally asymptotically stable if the following inequality holds for all {\small $n\in \mathbb{N}$}}
\vspace{-0.05in}
\begin{equation}
\small 
a_{d,n} <1.
\vspace{-0.03in}
\end{equation}
\label{Theorem1_SuffStabCond}
\vspace{-0.2in}
\end{thm}

The proof is in Section II-A of the supplementary file.

\subsubsection{Stability condition on footstep control}
Based on Theorem~\ref{Theorem1_SuffStabCond}, the following theorem provides the sufficient condition under which the footstep controller in \eqref{Eq-FeedbackSteppingControl}
asymptotically stabilizes the HT-LIP model in \eqref{Eq-HybridSys}.

\begin{thm}[\textbf{Sufficient stability condition on footstep control gain}]
%
%
{\it Consider assumptions (A1) and (A2).
The feedback footstep controller gain {\small $\mathbf{K}$} (i.e., {\small $k_1$} and {\small $k_2$}) guarantees the asymptotic closed-loop stability of the desired trajectory {\small $\mathbf{X}_r(t)$} for the HT-LIP model if }
\vspace{-0.05in}
\begin{equation}
   \small
   \begin{aligned}
       & \Bigg| (1- k_{1}) \cosh{(\xi_n)} \Bigg| + \Bigg| \tfrac{\sinh{(\xi_n)}}{\sqrt{\overline{f}_n}} -  k_{2}\cosh{(\xi_n)} \Bigg|
       < 1~\text{and}~
        \\
       & \Bigg|  (1- k_{1}) \sqrt{\overline{f}_n}\sinh{(\xi_n )}   \Bigg|
      + \Bigg|  \cosh{(\xi_n)}  - k_{2}\sqrt{\overline{f}_n}\sinh{(\xi_n )}   \Bigg|<1
      \label{Eq-Condition on GainTerms}
    \end{aligned}
    \vspace{-0.03in}
\end{equation}
{\it hold for
any {\small $n^{\text{th}}$} gait cycle ({\small $n \in \mathbb{N}$}).
Here, {\small$\xi_n :=\Delta\tau_n \sqrt{ \overline{f}_n}$}.}
\label{Theorem2_SuffStabCond_on_ContGain}
\end{thm}
\noindent \textit{Proof:}
The rationale of the proof is to show if \eqref{Eq-Condition on GainTerms} is valid for all {\small $n\in \mathbb{N}$} 
then the stability condition in Theorem~\ref{Theorem1_SuffStabCond} holds. 


By definition, the state-transition matrix {\small $\overline{\boldsymbol{\Phi}}(\overline{f}_n;\Delta \tau_n,0)$} for the state-space representation of the time-invariant supremum model in~\eqref{eq: supreme}
is given as:
\begin{equation}
\vspace{-0.05in}
   \small
   \begin{aligned}
    \overline{\boldsymbol{\Phi}}(\overline{f}_n;\Delta \tau_n,0)
    &=
    \text{exp} \big( {\begin{bmatrix}
        0 &1\\
        \overline{f}_n &0
    \end{bmatrix}\Delta \tau_n} \big)
    =:
\begin{bmatrix}
    \overline{\boldsymbol{\Phi}}_{11} & \overline{\boldsymbol{\Phi}}_{12} \\
    \overline{\boldsymbol{\Phi}}_{21} & \overline{\boldsymbol{\Phi}}_{22}
    \end{bmatrix}
    \\
    &
    =:
    \begin{bmatrix}
    &\cosh{(\xi_n )} 
    &\frac{\sinh{(\xi_n)}}{{\sqrt{\overline{f}_n}}}\\
    &\sqrt{\overline{f}_n}\sinh{(\xi_n )}
    &\cosh{(\xi_n)}
    \end{bmatrix}.
    \label{Eq-STM_SupLTI}
    \end{aligned}
\end{equation}

Using the expressions of the state-transition matrix in \eqref{Eq-STM_SupLTI} and those of {\small $\boldsymbol{\beta}$ and $\mathbf{K}$}, we can express {\small $\overline{\mathbf{A}}_{d,n}$} as: 
\vspace{-0.05in}
\begin{equation}
\small
   \begin{aligned}
   \overline{\mathbf{A}}_{d,n}
        =\begin{bmatrix}
       (1- k_{1})\overline{\boldsymbol{\Phi}}_{11} &   \overline{\boldsymbol{\Phi}}_{12}- k_{2}\overline{\boldsymbol{\Phi}}_{11}\\
       (1- k_{1})\overline{\boldsymbol{\Phi}}_{21} &\overline{\boldsymbol{\Phi}}_{22}- k_{2}\overline{\boldsymbol{\Phi}}_{21}
       \end{bmatrix}.
    \label{Eq-Adn_asFuncofK}
    \vspace{-0.03in}
\end{aligned}
\end{equation}

By definition, the infinity norm of {\small $\overline{\mathbf{A}}_{d,n}$} is:
\vspace{-0.05in}
\begin{equation}
\small
    \begin{aligned}
       \|\overline{\mathbf{A}}_{d,n} \| _\infty :=\max(& |\overline{\boldsymbol{\Phi}}_{11}(1- k_{1})| + |\overline{\boldsymbol{\Phi}}_{12}- \overline{\boldsymbol{\Phi}}_{11}k_{2}|,\\
       &|\overline{\boldsymbol{\Phi}}_{21}(1- k_{1})|+ |\overline{\boldsymbol{\Phi}}_{22}- \overline{\boldsymbol{\Phi}}_{21}k_{2}|).
       \label{Eq_Adn_infty}
    \end{aligned}
    \vspace{-0.03in}
\end{equation} 


If the footstep controller satisfies \eqref{Eq-Condition on GainTerms},
then {\small $|\overline{\boldsymbol{\Phi}}_{11}(1- k_{1})| + |\overline{\boldsymbol{\Phi}}_{12}- \overline{\boldsymbol{\Phi}}_{11}k_{2}| <1$ and $|\overline{\boldsymbol{\Phi}}_{21}(1- k_{1})|+ |\overline{\boldsymbol{\Phi}}_{22}- \overline{\boldsymbol{\Phi}}_{21}k_{2}|<1$} hold for any {\small $n \in \mathbb{N}$}.
Accordingly, {\small $a_{d,n}=\|\overline{\mathbf{A}}_{d,n} \| _\infty < 1$} holds on {\small $n \in \mathbb{N}$},
meeting the stability condition in Theorem \ref{Theorem1_SuffStabCond}.
$\hfill
\blacksquare$

\begin{rem}[\textbf{Applicability of Theorems \ref{Theorem1_SuffStabCond} and \ref{Theorem2_SuffStabCond_on_ContGain}}]
The stability conditions in Theorems \ref{Theorem1_SuffStabCond} and \ref{Theorem2_SuffStabCond_on_ContGain} are valid for a variable continuous-phase duration and a general (periodic and aperiodic) vertical DRS motion.
Also, applying these conditions does not require an accurate knowledge of the vertical DRS motion but an upper bound of its acceleration.
\end{rem}

\vspace{-0.15 in}
\subsection{Formulation of QP-based Footstep Control}
\vspace{-0.05 in}

To ensure online footstep planning,
we formulate a computationally efficient QP that calculates the controller gain {\small $\mathbf{K}$} in real-time, maximizes the error convergence rate, and enforces feasibility and stability conditions of the HT-LIP. 


\subsubsection{Ensuring real-time update of control gain}
Because the stability condition in Theorem~\ref{Theorem2_SuffStabCond_on_ContGain} relies on the values of 
the system parameter {\small $\xi_n$}
that can vary across different gait cycles, 
it is necessary to update the control gain {\small $\mathbf{K}$} {at least once per gait cycle} in order to meet the stability condition.
The variance of {\small $\xi_n$} across different gait cycles is due to changes in the gait cycle duration {\small $\Delta \tau_n$} and the parameter {\small $\overline{f}_n$}.
The varying value of {\small $\Delta \tau_n$} across gait cycles can be induced by users or a high-level path planner, while that of {\small $\overline{f}_n$} can be caused by the constantly changing DRS motion.


For timely mitigation of uncertainties in real-world applications, updating the planned footstep position every time step is necessary~\cite{gong2020angular}.
Although Theorem~\ref{Theorem2_SuffStabCond_on_ContGain} ensures the system stability under the once-per-gait-cycle update of {\small $\mathbf{K}$ and $\xi_n$}
instead of an update every time step,
Theorem~\ref{Theorem2_SuffStabCond_on_ContGain} can be readily extended to guarantee the stability even when {\small $\mathbf{K}$} and {\small $\xi_n$} are updated every time step.
This is essentially because the supremum system used to construct the stability conditions is time-invariant and accordingly 
its S2S state-transition matrix {\small $\overline{\mathbf{A}}_{d,n}$}
enjoys the associative property in terms of time {\small $t$} within each continuous phase.



\subsubsection{Ensuring fast convergence rate}

Lemma 1 in Sec. II-A-3) of the supplementary file shows that for all {\small $n\in \mathbb{N}$} we have
{\small $\Big\|\mathbf{e}|_{n+1}^- \Big\|\leq  
      {a_{d,n}}
      \Big \|\mathbf{e}|_n^-\Big\|$}.
Thus, minimizing {\small $a_{d,n}$} ensures a fast convergence rate of the error {\small $\mathbf{e}$}.
Based on~\eqref{Eq_Adn_infty}, this can be achieved by minimizing the sum of the squares of {\small $|\overline{\boldsymbol{\Phi}}_{11}(1- k_{1})| + |\overline{\boldsymbol{\Phi}}_{12}- \overline{\boldsymbol{\Phi}}_{11}k_{2}|$} and {\small $|\overline{\boldsymbol{\Phi}}_{21}(1- k_{1})|+ |\overline{\boldsymbol{\Phi}}_{22}- \overline{\boldsymbol{\Phi}}_{21}k_{2}|)$},
which is used as the cost function {\small $J(\mathbf{K})$}:
\vspace{-0.05in}
\begin{equation}
\small
    J(\mathbf{K}) = \frac{1}{2} \mathbf{KSK}^T +\mathbf{Kc}.
    \label{Eq_StepContGainCost}
    \vspace{-0.03in}
\end{equation}
Here {\small $\mathbf{S}$} and {\small $\mathbf{c}$} are respectively the Hessian matrix and gradient vector of the cost function {\small $J(\mathbf{K})$} and are defined as:
\vspace{-0.05in}
\begin{equation}
\small
    \begin{aligned}
    \small
        \mathbf{S} &=\begin{bmatrix} 2(\overline{\boldsymbol{\Phi}}_{11}^2+\overline{\boldsymbol{\Phi}}_{21}^2) &0\\
        0 & 2(\overline{\boldsymbol{\Phi}}_{11}^2+\overline{\boldsymbol{\Phi}}_{21}^2)    
        \end{bmatrix}~\text{and}~ \\
        \mathbf{c} &= [-2(\overline{\boldsymbol{\Phi}}_{11}^2+\overline{\boldsymbol{\Phi}}_{21}^2),~ -2(\overline{\boldsymbol{\Phi}}_{11}\overline{\boldsymbol{\Phi}}_{12}+\overline{\boldsymbol{\Phi}}_{21}\overline{\boldsymbol{\Phi}}_{22})]^T.
    \end{aligned}
    \label{Eq_HessianGradForHTLIP_QP}
    \vspace{-0.03in}
\end{equation}

\subsubsection{Enforcing stability conditions}
The asymptotic stability condition of the HT-LIP model under the proposed footstep control law, given in \eqref{Eq-Condition on GainTerms}, can be rewritten as:
\vspace{-0.05in}
\begin{equation}
\small
    \begin{aligned}          
                 \begin{bmatrix}
        -\overline{\boldsymbol{\Phi}}_{11} 
        &-\overline{\boldsymbol{\Phi}}_{11}
        \\
        ~~\overline{\boldsymbol{\Phi}}_{11} 
        &~~\overline{\boldsymbol{\Phi}}_{11}
        \\
        -\overline{\boldsymbol{\Phi}}_{21} 
        &-\overline{\boldsymbol{\Phi}}_{21}
        \\
        ~~\overline{\boldsymbol{\Phi}}_{21} 
        &~~\overline{\boldsymbol{\Phi}}_{21}
        \end{bmatrix}\mathbf{K} 
        < 
        \begin{bmatrix}
        1-\overline{\boldsymbol{\Phi}}_{11}-\overline{\boldsymbol{\Phi}}_{12} \\ 
        1+ \overline{\boldsymbol{\Phi}}_{11}+ \overline{\boldsymbol{\Phi}}_{12}\\ 
        1-\overline{\boldsymbol{\Phi}}_{21}-\overline{\boldsymbol{\Phi}}_{22} \\ 
        1+ \overline{\boldsymbol{\Phi}}_{21}+ \overline{\boldsymbol{\Phi}}_{22}  
        \end{bmatrix}.
    \end{aligned}
    \label{Eq_StabilityConstraint}
    \vspace{-0.03in}
\end{equation}

\subsubsection{Satisfying kinematic limits and ground-contact constraints}
The physical feasibility of footstep planning is guaranteed by respecting (i) the kinematic bounds on the trotting step length and (ii) the friction cone and unilateral ground-contact constraints.
The kinematic limit of the step length {\small $u_{x,d}$} can be expressed as {\small $u_{x,d}\in[u_{min},~u_{max}]$}, where {\small $u_{max}$} and {\small $u_{min}$} are the maximum and minimum step lengths of the HT-LIP, respectively.
Meanwhile, the step length should be set to respect the friction cone and unilateral constraints at the foot-surface contact points expressed as {\small $u_{x,d} \in [- 2\mu z_0, 2\mu z_0]$}, where {\small $\mu$} is the friction coefficient.


In summary, the stability condition and the feasibility constraints can be compactly expressed as:
\vspace{-0.05in}
\begin{equation}
\small
    \mathbf{EK}^T<\mathbf{d}
    \label{Eq_FootStepGainQPConst}
    \vspace{-0.05in}
\end{equation}
with
\vspace{-0.05in}
\begin{equation}
\small
    \begin{aligned}          
                 \mathbf{E} &:=\begin{bmatrix}
        ~~{e} & ~~ {\dot{e}}\\
        -{e} &-{\dot{e}}\\
        -\overline{\boldsymbol{\Phi}}_{11} 
        &-\overline{\boldsymbol{\Phi}}_{11}
        \\
        ~~\overline{\boldsymbol{\Phi}}_{11} 
        &~~\overline{\boldsymbol{\Phi}}_{11}
        \\
        -\overline{\boldsymbol{\Phi}}_{21} 
        &-\overline{\boldsymbol{\Phi}}_{21}
        \\
       ~~ \overline{\boldsymbol{\Phi}}_{21} 
        &~~\overline{\boldsymbol{\Phi}}_{21}
        \end{bmatrix}~\text{and}~ 
        \mathbf{d} := 
        \begin{bmatrix}
        l_{max}-u_{r,n}\\
        -l_{min}+u_{r,n}\\
        1-\overline{\boldsymbol{\Phi}}_{11}-\overline{\boldsymbol{\Phi}}_{12} \\ 
        1+ \overline{\boldsymbol{\Phi}}_{11}+ \overline{\boldsymbol{\Phi}}_{12}\\ 
        1-\overline{\boldsymbol{\Phi}}_{21}-\overline{\boldsymbol{\Phi}}_{22} \\ 
        1+ \overline{\boldsymbol{\Phi}}_{21}+ \overline{\boldsymbol{\Phi}}_{22}     
        \end{bmatrix},
    \end{aligned}
    \label{Eq_StepContConstDetails}
    \vspace{-0.05in}
\end{equation}
where the scalar, real constants {\small $l_{max}$} and {\small $l_{min}$} are defined as {\small $l_{max}:= \max(u_{max}, ~\mu z_0)$} and {\small $l_{min}:= \min(u_{min},~ -\mu z_0)$}.

With the cost function and constraints designed, the proposed QP that produces the footstep controller gain {\small $\mathbf{K}$} is given in the following theorem.

\begin{thm}[\textbf{QP-based control gain optimization}]
{\it The control gain {\small $\mathbf{K}$} that maximizes the convergence rate, guarantees stability, and ensures feasibility for an HT-LIP model is given as a solution to the following QP problem: }
\vspace{-0.05in}
\begin{equation}
\small
\begin{aligned}
\min_{\mathbf{K}} \quad & J(\mathbf{K}) \\
\text{subject to} \quad & \mathbf{EK}^T<\mathbf{d}.
\label{Eq_QP_SteppingGain}
\end{aligned}  
\vspace{-0.05in}
\end{equation}
\label{Theorem3_QP_basedContGain}
\end{thm}

The proof is given in Sec. II-B of the supplementary file.


\begin{rem}[\textbf{Solution feasibility and optimality of the proposed QP}]
Note that the cost function in \eqref{Eq_StepContGainCost} 
is convex.
Meanwhile, the feasibility and stability constraints of the QP in \eqref{Eq_QP_SteppingGain} are non-conflicting if the feasible region for the constraints {\small $\mathbf{EK}^T<\mathbf{d}$} remains non-empty.
Accordingly, the solution feasibility and optimality for the QP problem in \eqref{Eq_QP_SteppingGain} is guaranteed.
In practice, the non-emptiness of the feasible region can be numerically evaluated under the admissible range of system parameters {\small $\overline{f}_n$} and {\small $\Delta \tau_n$}.
\label{Remark2_FeasOptSol}   
\end{rem}


\begin{rem}[\textbf{Solving the QP in real-time}]
Solving the proposed QP requires the knowledge of the upper bound of {\small $f(t)$} during any {\small $n ^{\text{th}}$} gait cycle, as indicated by the stability condition in Theorem~\ref{Theorem2_SuffStabCond_on_ContGain}.
Since the needed upper bound can be any upper bound of {\small $f(t)$} during any {\small $n ^{\text{th}}$} gait cycle, we can solve the proposed QP, in principle, by using a sufficiently large value of the upper bound {\small $\overline{f}_n$} that is valid across any {\small $n ^{\text{th}}$} gait cycles.
Yet, using such a bound might be overly conservative, reducing locomotion robustness.
Thus, we choose to estimate the upper bound of the surface acceleration in real-time and update {\small $\overline{f}_n$} at every time step.

The vertical surface acceleration {\small $\ddot{z}_s$} can be roughly estimated based on the readings of an on-board inertial measurement unit (typically placed at the trunk) and the robot's forward kinematics.
Using the rough estimate, we can then obtain both an upper bound of the surface acceleration {\small $\ddot{z}_s$} and the values of {\small $\overline{f}_n$}, {\small $a_{d,n}$}, and {\small $\overline{\boldsymbol{\Phi}}$}.
\end{rem}


\vspace{-0.1 in}
\section{Experiments} 
\vspace{-0.05 in}
\label{Section-Experiments}

This section presents hardware experiment results to demonstrate the proposed control framework can stabilize quadrupedal trotting on a DRS with an aperiodic and unknown vertical motion even in the presence of various uncertainties.
The experiment video is in a supplementary file and is also available at \href{https://youtu.be/BMPU0BJQC64}{https://youtu.be/BMPU0BJQC64}.

\vspace{-0.15 in}
\subsection{Hardware Experiment Setup}
\vspace{-0.05 in}

\subsubsection{Treadmill} 
\begin{figure}[t]
    \centering
    \includegraphics[width= 1\linewidth]{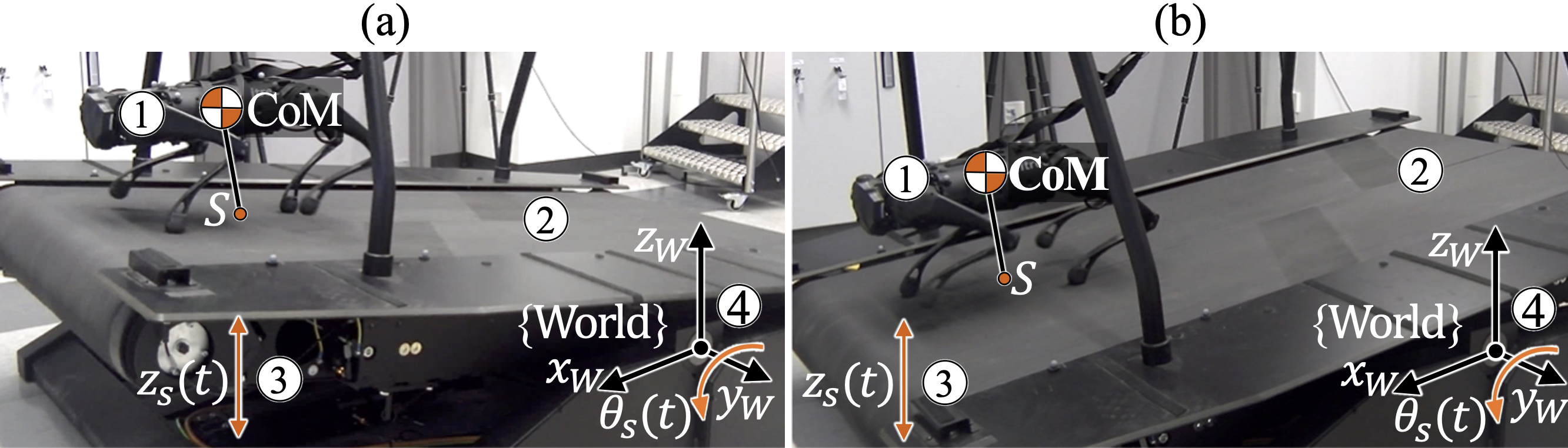}
    \vspace{-0.25 in}
    \caption{Illustration of the experimental setup. \raisebox{.5pt}{\textcircled{\raisebox{-.9pt} {1}}}: Go1 quadruped (Unitree Robotics). \raisebox{.5pt}{\textcircled{\raisebox{-.9pt} {2}}}: M-Gait treadmill (Motek Medical). \raisebox{.5pt}{\textcircled{\raisebox{-.9pt} {3}}}: direction of the vertical DRS/treadmill motion $z_s(t)$ at point $S$. \raisebox{.5pt}{\textcircled{\raisebox{-.9pt} {4}}}: world frame attached to the treadmill's axis of pitching. The treadmill's pitch angle at time $t$ is $\theta_s(t)$. Subplots (a) and (b) show the treadmill at its pitch angle limits.}
    \label{Fig:Go1_ExpSetup}
        \vspace{-0.15 in}
\end{figure}

Our experiments use a Motek M-Gait treadmill to emulate a vertically moving DRS (Fig.~\ref{Fig:Go1_ExpSetup}).
The treadmill can perform pre-programmed pitch and sway movements.
It weighs {\small 750} kg, measures {\small 2.3} m $\times$ {\small 1.82} m $\times$ {\small 0.5} m, and is equipped with two belts (each powered by a {\small 4.5} kW servo motor).
The robot is positioned approximately {\small 0.8} m from the treadmill's pitching axis.

\subsubsection{Unknown vertical treadmill/DRS motions}
\label{Sec 5: vertical motion}
The experiments utilize the treadmill's pitch motion {\small $\theta_s(t)$} to generate aperiodic, vertical DRS motions at the robot's footholds (i.e., near the treadmill's far end). 
Table \ref{Table_HardwareCases} summarizes the surface motions (HC1)-(HC5), which are unknown to the proposed control framework during experiments.
Although the pitch angle {\small $\theta_s(t)$} is small, it induces a significant maximum vertical acceleration {\small $\ddot{z}_s(t)$} at the robot's footholds (about {\small $3.5$} m/s$^2$) with a minimal horizontal surface motion.
Figures 1-4 in the supplementary file illustrate (HC1)-(HC3) and (HC5).

\begin{table}[t]
    \centering
    \caption{DRS motions under different hardware experiment cases.}
    \vspace{-0.1 in}
 \begin{tabularx}{0.9\columnwidth}{c|c}
 \hline
 \hline
 {\small Cases} & ~~~~~~~~~~~~~{\small DRS motion}\\
 \hline
 \hline
 {\small (HC1)}   & {\small $\theta_s(t) = 4^\circ(\sin3t+ \sin(t\sqrt{0.5t + 1})) .$} \\
  \hline
{\small (HC2)} &   {\small $\theta_s(t) = 4^\circ(\sin6t+ \sin (0.1t^2)).$} \\
  \hline
{\small (HC3)} & {\small $\theta_s(t) = 0.2^\circ t^2\sin\left(\sqrt{100t+1}\right) \cdot e^{-t/10}.$}
\\
  \hline
{\small (HC4)}    & {\small $\theta_s(t) = 4^\circ(\sin3t+ \sin(t\sqrt{t/2 + 1}))$}~{\small \text{and}}~ 
\\   
~~~  & {\small $y_s(t) = 
\begin{cases}
0,~\text{if}~0~\text{s}\leq t \leq 83 ~\text{s};\\
40 \sin (\pi t)~\text{mm},~\text{if}~83~\text{s}<t\leq 122~\text{s};
\\   
65 \sin (\pi t)~\text{mm},~\text{if}~122~\text{s}<t\leq 160~\text{s}.
\end{cases}
$} 
\\
 \hline
 {\small (HC5)}  & {\small $\theta_s(t) = 2.5^\circ(\sin3t+ \sin(t \sqrt{0.5t + 1})).$ } 
 \\
 \hline
\end{tabularx}
\label{Table_HardwareCases}
\vspace{-0.2in}
\end{table}



%
\subsubsection{Additional uncertainties}
\label{Sec 5: uncertain}
To validate the robustness of the proposed approach beyond unknown vertical DRS motions, we test additional unmodeled uncertainties (Fig.~\ref{Fig:HardwareValidation}).

\begin{figure}[t]
    \centering
    \includegraphics[width= 0.95\linewidth]{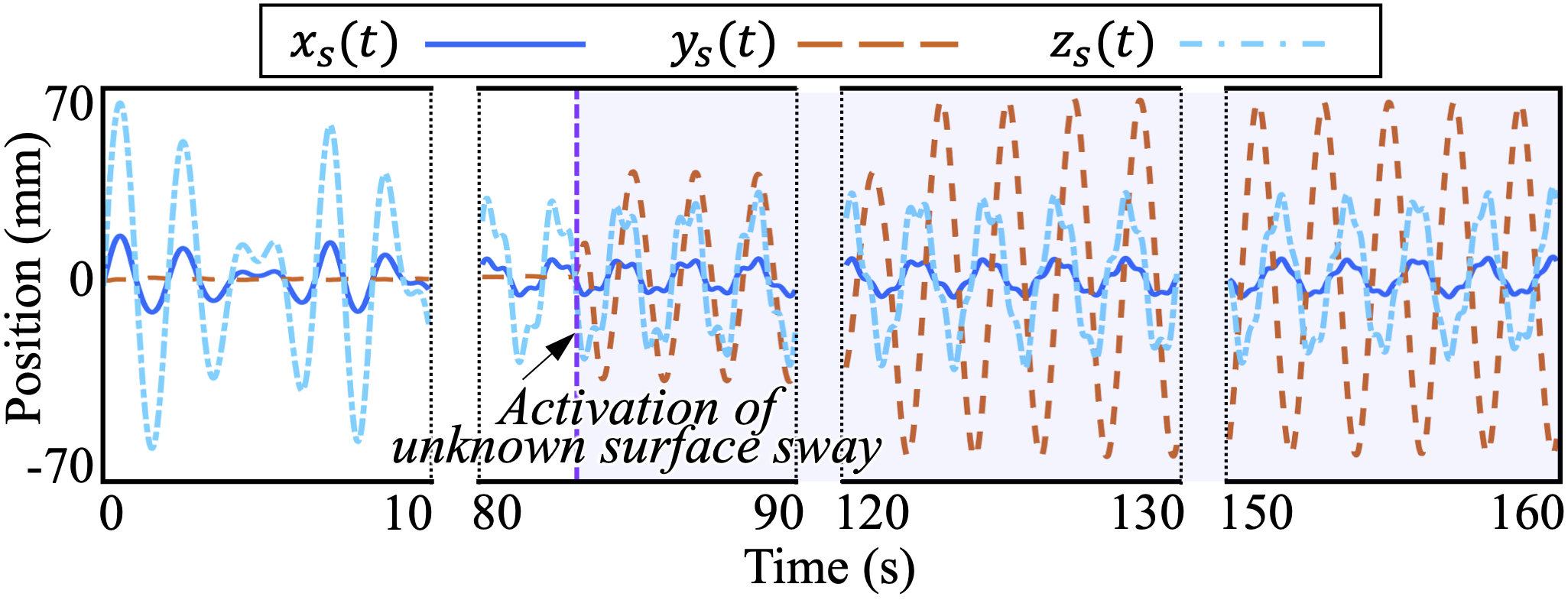}
    \vspace{-0.15 in}
    \caption{Ground-truth position trajectory of the point on the treadmill/DRS around which the robot performs the trotting gait during the unknown pitch and sway movement (HC4) of the DRS.
    The shaded area highlights the period during which the unknown DRS sway motion is active.}
\label{Fig:DRSCase4}
    \vspace{-0.15 in}
\end{figure}

To assess the robustness against unknown DRS sway, the surface motion (HC4) contains a sway displacement {\small $y_s(t)$} (see Table~\ref{Table_HardwareCases} and Fig.~\ref{Fig:DRSCase4}), causing a peak horizontal acceleration of {\small $2.6$} m/s$^2$ at the robot's footholds.

Besides surface sway, four other types of uncertainties are tested during (HC5) with
maximum vertical and lateral accelerations respectively at {\small $1.5$} m/s$^2$ and {\small $0.5$} m/s$^2$. 
These uncertainties are: (i) uncertain friction coefficient of {\small 0.3}-{\small 0.4} induced by a smooth glass surface while the framework considers a coefficient of {\small 0.8}; (ii) unknown solid ({\small 10} lbs) and liquid ({\small 9} lbs) loads placed on the trunk， weighing respectively {\small 36}\% and {\small 32}\% of the robot's mass; (iii) uneven (pebbled) surface with a maximum height of {\small 10} cm; and (iv) sudden pushes
lasting less than {\small 0.2} s per push and inducing a robot heading error of {\small 30}$^\circ$ just after the push.

\vspace{-0.15 in}
\subsection{Control framework setup}
\vspace{-0.05 in}

The HT-LIP model parameters considered by the proposed control framework are given in Table~\ref{table:H-LIP_Param}.
These parameters are varied during experiments to demonstrate the control framework can be implemented in real-time under different trotting gait features.
The framework explicitly considers the vertical DRS acceleration {\small $\ddot{z}_s(t)$} and assumes negligible horizontal DRS motion, and only considers the estimated instead of the true value of {\small $\ddot{z}_s(t)$}.
With the estimation method mentioned in Remark 3, the maximum absolute error of the vertical DRS motion estimation is {\small $1$} m/s${}^2$.

\begin{table}[t]
    \centering
    \caption{Ranges of HT-LIP parameters used in experiments}
    \label{table:H-LIP_Param}
    \vspace{-0.1 in}
    \begin{tabularx}{0.85\columnwidth}{c|c}
     \hline
      \hline
     {\small Parameter}    &  {\small Range}
    \\
     \hline
    \hline
    {\small CoM height above the surface} {\small $z_0$} (cm) 
       & {\small $[22,~26]$}
     \\
     {\small Step duration} {\small $\Delta \tau_n$} (s)  &  {\small $[0.15,~0.4]$}    
     \\
     {\small Walking speed (cm/s)}   &  {\small $[15,~25]$}
     \\
     {\small Nominal step length} {\small $u_{x,r}$} {\small(cm)}   & {\small $[0,~15]$} 
     \\
     \hline
     \hline
    \end{tabularx}
    \vspace{-0.2in}
\end{table}

\vspace{-0.15 in}
\subsection{Experimental Results}
\vspace{-0.05 in}

This subsection reports the experiment results under unknown DRS motions and various other types of uncertainties.

\subsubsection{Validation under unknown vertical surface motions}

\begin{figure}[t]
    \centering
    \includegraphics[width= 0.9\linewidth]{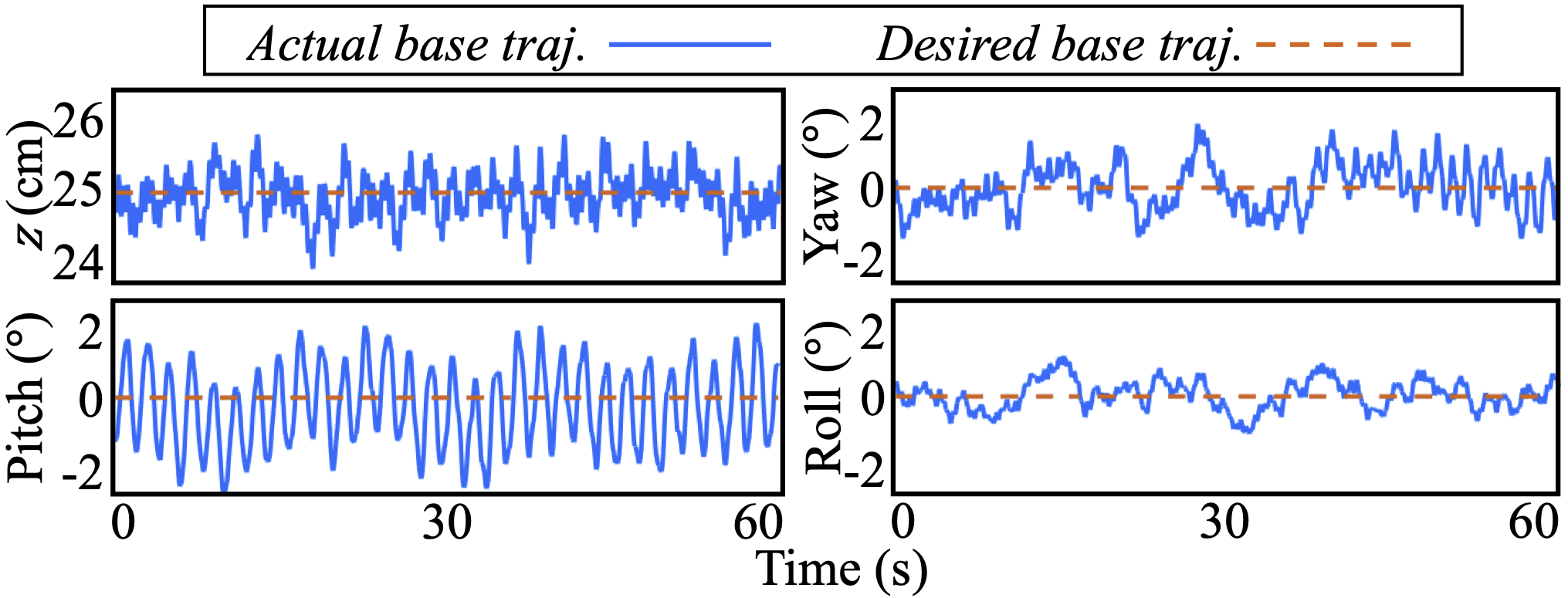}
    \vspace{-0.15 in}
    \caption{Desired and actual base trajectories under the hardware experiment case (HC1). The small tracking errors indicate stable robot trotting.}
    \label{Fig:DRSCase1r}
    \vspace{-0.1 in}
\end{figure}

\begin{figure}[t]
    \centering
    \includegraphics[width=0.95\linewidth]{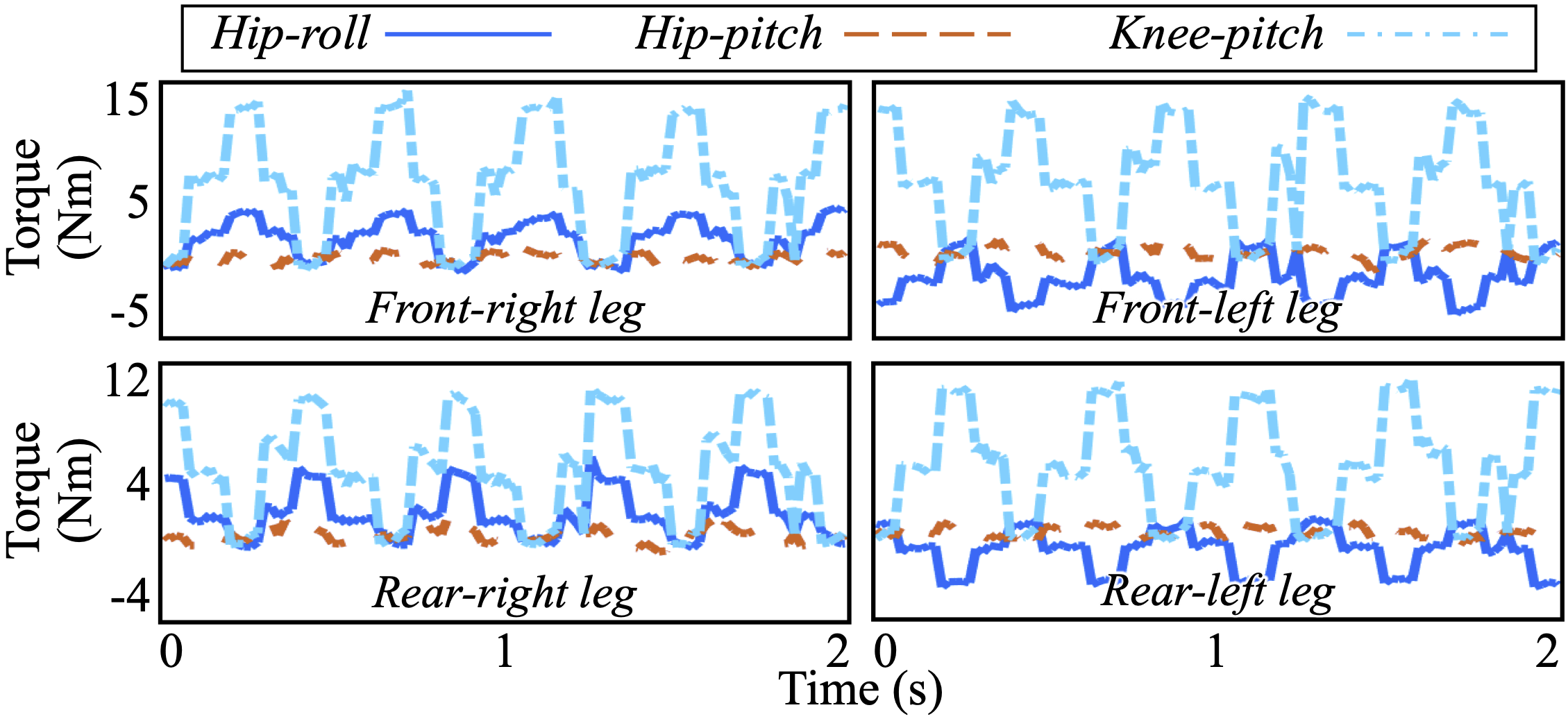}
    \vspace{-0.15in}
    \caption{Torque profiles under the hardware experiment case (HC1), all of which respect the robot's individual actuator limit of 22.5 Nm.}
    \label{Fig:Case1_JT}
    \vspace{-0.2 in}
\end{figure}

As shown in Fig.~\ref{Fig:DRSCase1r}, the actual height and orientation of the robot's base (i.e., trunk) relatively closely track the desired base trajectories during the unknown and aperiodic vertical surface motion (HC1), indicating a stable trotting gait under the proposed control framework. 
Further, the joint torque profile in Fig.~\ref{Fig:Case1_JT} demonstrates a consistent torque pattern that respects the actuator limit {of {\small 22.5} N/m for all joints}.

From Figs. 5-8 in the supplementary file, results under (HC2) and (HC3) also show accurate trajectory tracking and consistent torque profiles, highlighting the effectiveness of the framework in handling different vertical DRS motions. 

\begin{figure}[t]
    \centering
    \includegraphics[width= 1\linewidth]{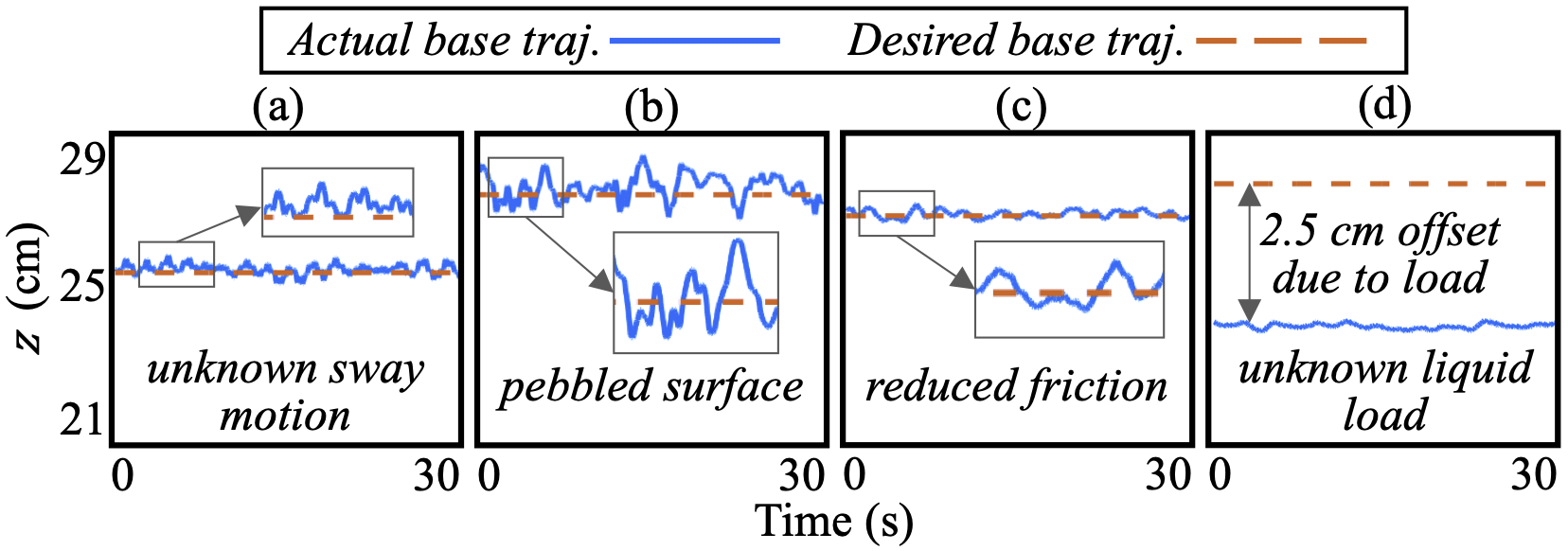}
    \vspace{-0.33 in}
    \caption{Base height trajectories under various cases of uncertainties, all during the unknown vertical DRS motion (HC5). 
    These cases include (a) unknown sway motion, (b) pebbled surface with an unknown height, (c) surface with unknown reduced friction, and (d) unknown load. }
    \label{Fig:Robust_Base_Height}
    \vspace{-0.1 in}
\end{figure}

\subsubsection{Validation under various additional uncertainties}
\label{Sec-ValidationUnderUncertainty}

To further assess robustness, we conduct hardware experiments under uncertainty cases described in Sec.~\ref{Sec 5: uncertain}.

The subplots (a) and (c) in Fig.~\ref{Fig:Robust_Base_Height} confirm that the robot's base height closely follows the desired value even under the unknown DRS sway motion and reduced surface friction.
The subplot (b) shows a notable oscillatory deviation of the actual base height from the desired value due to the unevenness of the pebbled surface, indicating a moderate level of violation of the constant base height assumption (i.e. assumption (A3)).
The subplot (d) shows the significant uncertain liquid load 
applied to the robot's trunk causes a nearly constant base height tracking error of {\small $2.5$} cm.
Still, both subplots (b) and (d) indicate stable locomotion despite uncertainties.
The results under the unknown solid load are similar to subplot (d) and thus are omitted for brevity.

\begin{figure}[t]
    \centering
    \includegraphics[width= 0.92\linewidth]{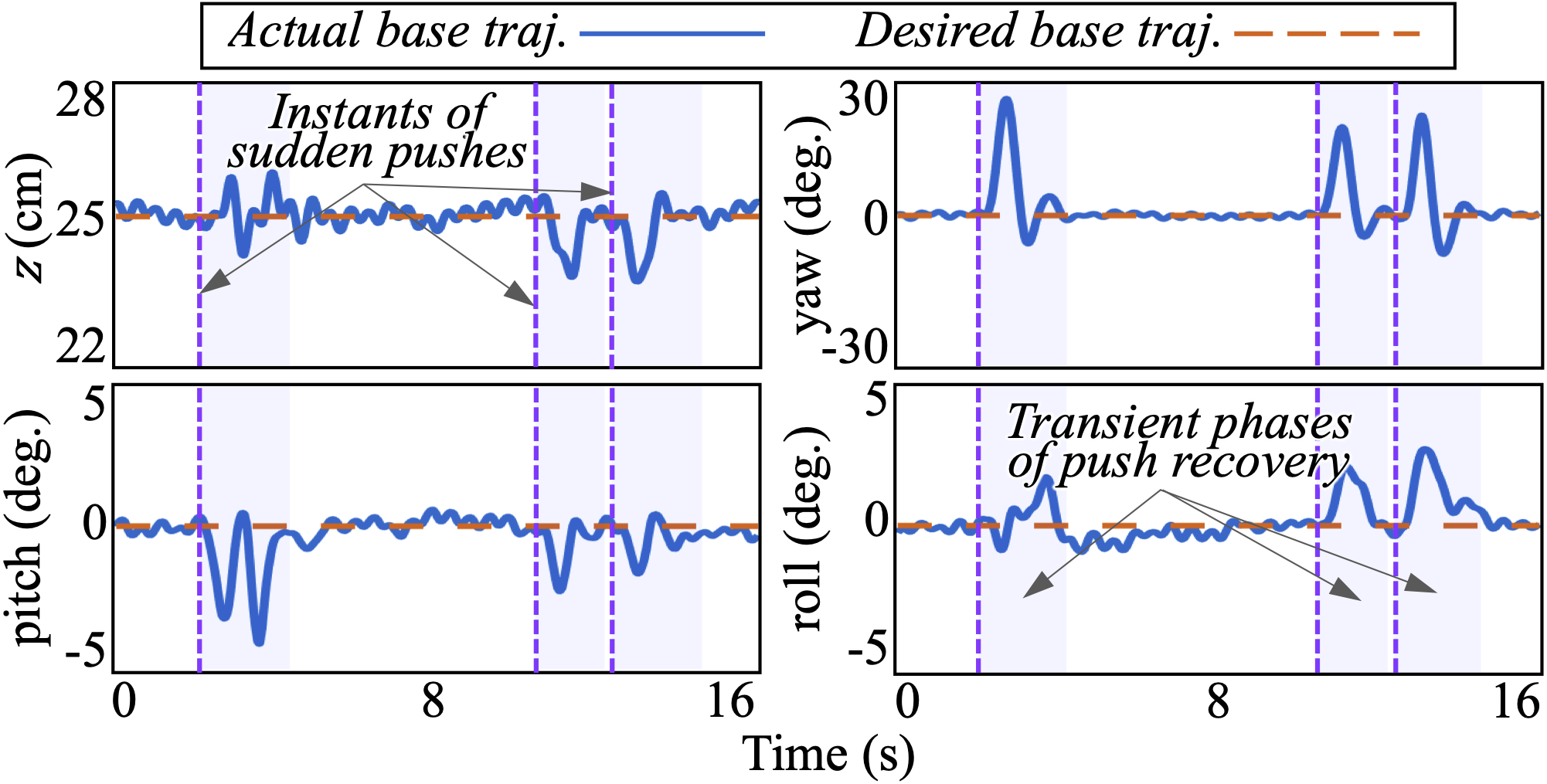}
    \vspace{-0.15 in}
    \caption{Robustness to sudden pushes under the uncertain DRS motion (HC4). 
    The purple dashed lines highlight the push instants, while the shaded regions show the transient push recovery phases. The proposed control framework effectively drives the perturbed trajectories to a close neighborhood of their desired values within 2 seconds.
    \vspace{-0.2 in}}
    \label{Fig:DRSCase4r_push}
\end{figure}

Figure~\ref{Fig:DRSCase4r_push} displays the push recovery results during the unknown vertical and lateral DRS motion (HC4).
The intermittent spikes in the robot's base height and orientation trajectories are induced by external pushes. %
As highlighted by the shaded areas in Fig.~\ref{Fig:DRSCase4r_push}, the robot is able to recover within two seconds after each significant push, confirming the robustness of the proposed framework against external pushes during unknown DRS motions.

\vspace{-0.15 in}
\subsection{Comparative Experiments}
\vspace{-0.05 in}

To show the improved robustness of our proposed framework compared to existing controllers, we experimentally test the Go1 robot's proprietary controller and a state-of-the-art baseline controller~\cite{cheetah3} during unknown vertical surface motion (HC5).
The baseline control approach has the same lower-layer torque controller as the proposed framework, but its higher and middle layers assume a static ground as designed in~\cite{cheetah3}.
Both the baseline and the proposed frameworks use the same filter introduced in~\cite{cheetah3} to estimate the robot's absolute base pose and velocity in real-time.

\begin{figure}[t]
    \centering
    \includegraphics[width= 1\linewidth]{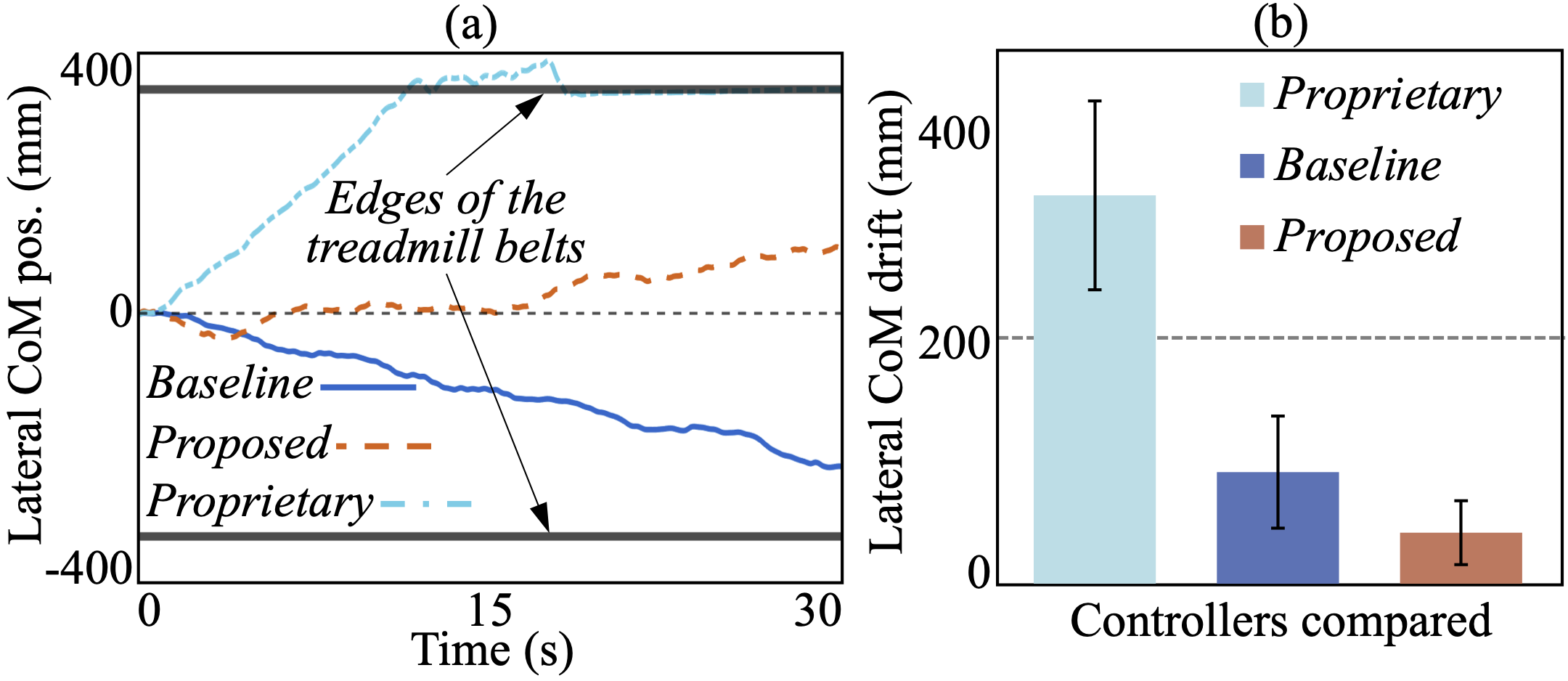}
    \vspace{-0.3 in}
    \caption{Lateral-position drift comparison with the robot's proprietary controller and a state-of-the-art controller~\cite{cheetah3} during the DRS motion (HC5): (a) lateral CoM position drift during a representative hardware experiment of 30 s and (b) average lateral drift (mean $\pm$ one standard deviation) during five experiment trials of 15 s.
    The proposed control approach achieves the least amount of lateral drift among the three approaches compared.} 
    \label{Fig:CompLatDrift}
    \vspace{-0.2 in}
\end{figure}

As illustrated by the lateral base position trajectory in Fig. \ref{Fig:CompLatDrift}, the proposed framework realizes the lowest lateral drift among the three approaches during trotting in place.
The relatively small lateral drift of the proposed framework is partly due to the explicit treatment of the unknown DRS motion in the higher-layer planner, which is missing in the baseline controller.
Also, both our framework and the baseline approach correct
the robot's heading direction
based on the estimated absolute base position and yaw angle.
In contrast, given the fast lateral position drift
under the proprietary controller, it is possible that the proprietary controller does not compensate for the base position error.

The proposed approach exhibits a lateral drift of approximately {\small 10} cm between {\small $t=15$} s and {\small $t=30$} s, mainly due to the drift of the estimated absolute base position and yaw angle of the robot~\cite{gao2022invariant}.
To improve its path tracking accuracy, 
a more accurate state estimator will be developed and used in our future work.
Note that this position drift is still notably lower than the drift under the baseline controller, which is over {\small 25} cm within {\small 30} seconds of trotting. Also, under the proprietary controller, the robot laterally drifts for approximately {\small 40} cm and hits the treadmill edge within the initial {\small 15} seconds.

\vspace{-0.15 in}
\section{Discussions}
\vspace{-0.05 in}
\label{Section-discussion}


%
One key contribution of this study is the introduction of the HT-LIP model for locomotion during general (periodic or aperiodic) and vertical DRS motions.
Similar to existing LIP models for static surfaces \cite{xiong2021_3d, dai2022bipedal,kajita20013d,caron2019capturability}, the HT-LIP model is linear.
Yet, the model is also explicitly time-varying due to the surface motion, distinguishing it from the time invariance of those existing models.
Meanwhile, since the model is homogeneous, it is fundamentally different from the LIP model for horizontally moving surfaces~\cite{gao2022time}.
Further, the HT-LIP is hybrid and is thus distinct from our previous continuous-time LIP model for vertical DRS motions~\cite{iqbal2022drs}.

Another key contribution is the construction of a discrete-time footstep controller that provably 
stabilizes the HT-LIP system under variable footstep duration and unknown vertical DRS motions.
The proposed stability condition for the footstep controller explicitly treats the time dependence of the HT-LIP model, which is fundamentally different from the previous footstep controller~\cite{xiong2021_3d} designed for static terrain.
Further, the proposed controller only consider a finite bound of the surface acceleration whereas our previous DRS locomotion controller~\cite{iqbal2020provably,iqbal2022drs} replies on an accurate knowledge of the surface motion.
Finally, the HT-LIP footstep controller is cast as a QP that enables real-time, feasible foot
placement while exactly enforcing the stability condition.

The experiments reveal that the proposed framework can handle a significant level of unknown DRS sway (up to {\small $2.6$} m/s${}^2$), although it does not explicitly treat unknown horizontal motions. Our future work will extend the proposed theoretical results and control framework from vertical DRS motions to simultaneous surface translation and rotation.

\vspace{-0.15 in}
\section{Conclusion}
\vspace{-0.05 in}
\label{section-conclusions}
%
This paper has introduced a hierarchical control framework for robust quadrupedal trotting 
during unknown and general vertical ground motions.
A reduced-order model was derived by analytically extending the existing linear, time-invariant H-LIP model to explicitly consider the surface motion, resulting in a hybrid, time-varying LIP model (i.e., HT-LIP).
Taking the HT-LIP as a basis, a discrete-time, provably stabilizing footstep controller was constructed and then cast as a quadratic program to enable real-time foot placement planning.
The proposed control framework incorporated the HT-LIP footstep controller as a higher-layer planner, and its middle and lower layers were developed to plan and control the robot's full-body motions that agree with the desired robot motions supplied by the higher layer.
Experiment results confirmed the robustness of the proposed framework in realizing stable quadrupedal trotting under various unknown, aperiodic surface motions, external pushes and (solid and liquid) loads, and slippery and rocky surfaces.

%
%


\vspace{-0.18 in}
\bibliography{IqbalTMECHAIM}
\bibliographystyle{ieeetr}

\vspace{-0.4 in}
\begin{IEEEbiography}[{\includegraphics[width=1in,height=1.15in,clip,keepaspectratio]{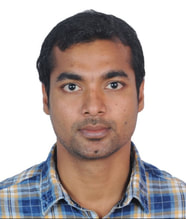}}]{Amir Iqbal}  is a Ph.D. candidate in Mechanical Engineering at the University of Massachusetts Lowell and a Research Intern at Purdue University. He received a B.Tech. degree in Aerospace Engineering from the Indian Institute of Space Science and Technology, Thiruvananthapuram, Kerala, India in 2012. He was a former Scientist/Engineer at the ISRO Satellite Center, Bengaluru, India. His research interests include legged locomotion, control theory, trajectory optimizations, and exoskeleton control.
\end{IEEEbiography}

\vspace{-0.5 in}
\begin{IEEEbiography}[{\includegraphics[width=1in,height=1.15in,clip,keepaspectratio]{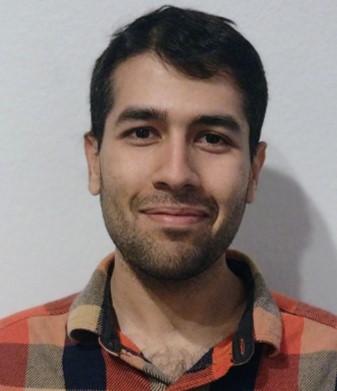}}]{Sushant Veer}
is a Senior Research Scientist at NVIDIA Research. In the past he was a Postdoctoral Research Associate in the Mechanical and Aerospace Engineering Department at Princeton University. He received his Ph.D. in Mechanical Engineering from the University of Delaware in 2018 and a B. Tech. in Mechanical Engineering from the Indian Institute of Technology Madras in 2013. His research interests lie at the intersection of control theory and machine learning with the goal of enabling safe decision making for robotic systems. He has received the Yeongchi Wu International Education Award (2013 International Society of Prosthetics and Orthotics World Congress), Singapore Technologies Scholarship (ST Engineering Pte Ltd), and Sri Chinmay Deodhar Prize (Indian Institute of Technology Madras).
\end{IEEEbiography}
\vspace{-0.4 in}
\begin{IEEEbiography}
[{\includegraphics[width=1in,height=1.15in,clip]{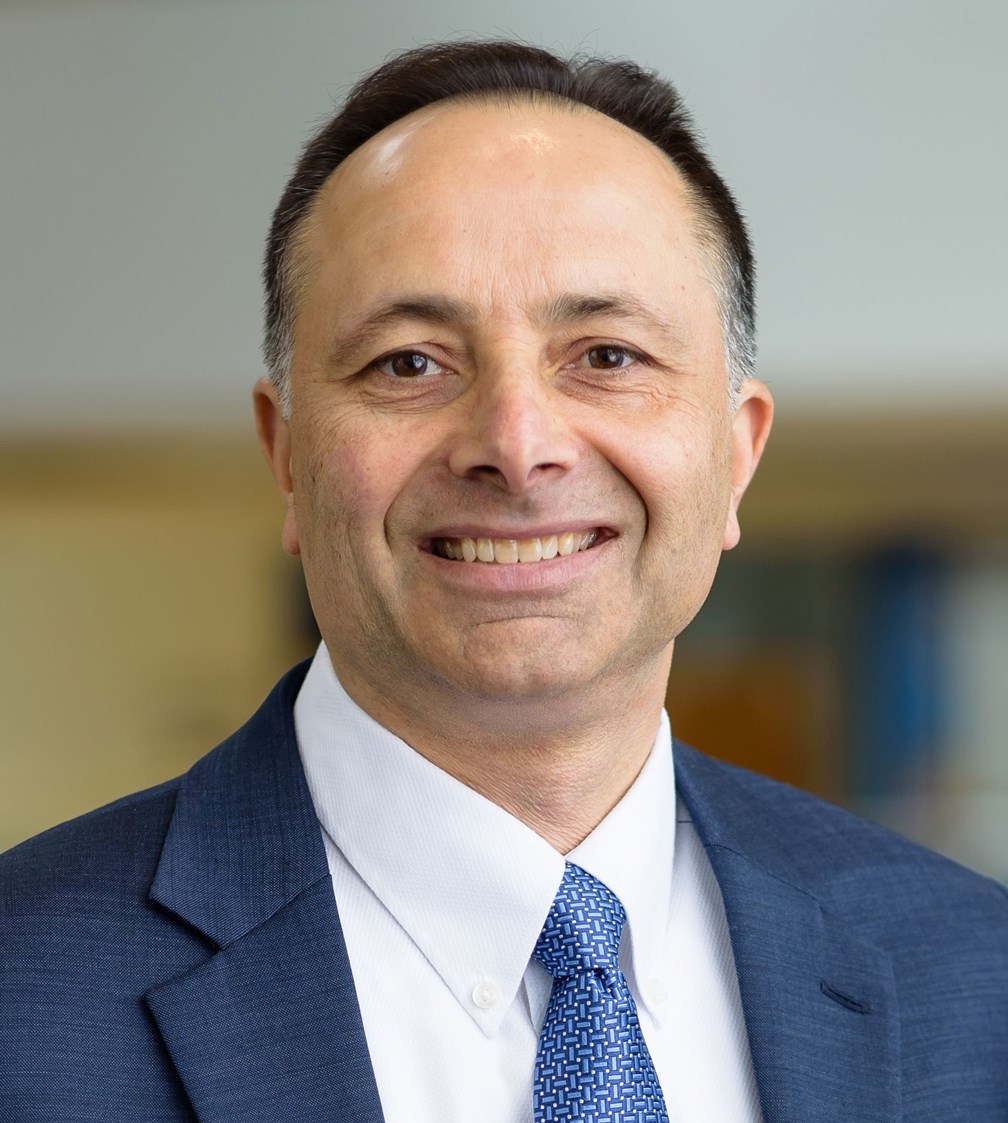}}]{Christopher Niezrecki}
received the dual B.S. degrees in mechanical and electrical engineering
from the University of Connecticut, Mansfield, CT, USA, in 1991, and the M.S. and Ph.D. degrees in mechanical engineering from Virginia Tech, Blacksburg, VA, USA, in 1992 and 1999, respectively. He is currently a Professor with the Department of Mechanical Engineering, University of
Massachusetts Lowell, Lowell, MA, USA, where he is also a Co-Director of the Structural Dynamics and Acoustics Systems Laboratory. He has been directly involved in mechanical design, smart structures, noise and vibration control, wind turbine blade dynamics and inspection, structural dynamic and acoustic systems, structural health monitoring, and noncontacting inspection research for over 30 years with more than 190 publications.
\end{IEEEbiography}

\vspace{-0.4 in}
\begin{IEEEbiography}
[{\includegraphics[width=1in,height=1.15in,clip]{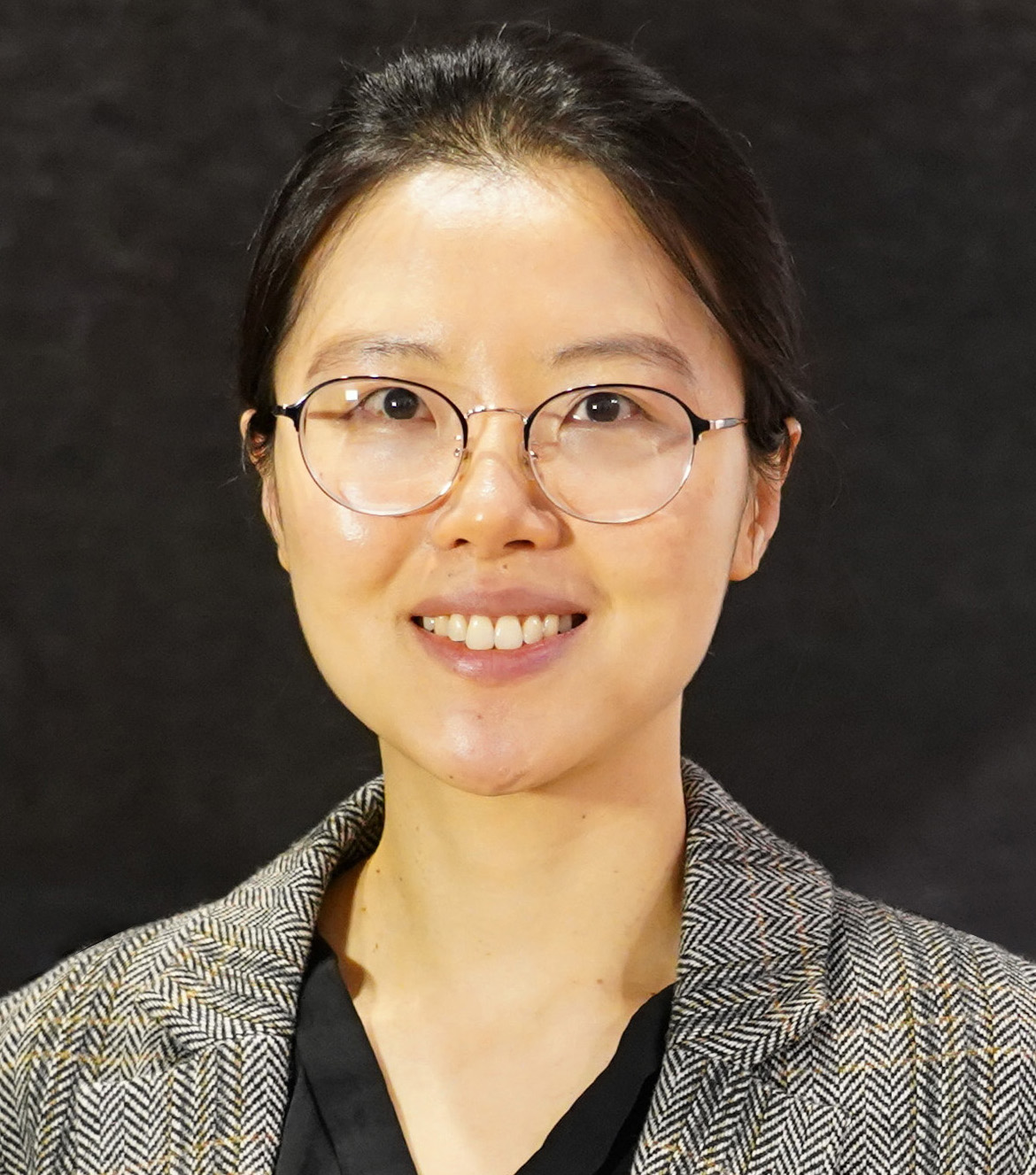}}]{Yan Gu}
received the B.S. degree in Mechanical Engineering from Zhejiang University, China, in June 2011 and the Ph.D. degree in Mechanical Engineering from Purdue University, West Lafayette, IN, USA, in August 2017.
She joined the faculty of the School of Mechanical Engineering at Purdue University in July 2022.
Prior to joining Purdue, she was an Assistant Professor with the Department of Mechanical Engineering at the University of Massachusetts Lowell.
Her research interests include nonlinear control, hybrid systems, legged locomotion, and wearable robots.
She was the recipient of the National Science Foundation Faculty Early Career Development Program (CAREER) Award in 2021 and the Office of Naval Research Young Investigator Program (YIP) Award in 2023.
\end{IEEEbiography}



\ifCLASSOPTIONcaptionsoff
  \newpage
\fi

\end{document}